   \documentclass[11pt|a4paper]{article}
 \usepackage[a4paper]{geometry}
\usepackage{acl2014}
\usepackage{times}
\usepackage{url}
\usepackage{amsmath}
\usepackage{latexsym}
\usepackage{natbib}
\usepackage{marginnote}
\usepackage{epsfig}
\usepackage{todonotes}
\usepackage[ruled,vlined]{algorithm2e}

\usepackage{times}
\usepackage{url}
\usepackage{latexsym}

\title{Classifying Idiomatic and Literal Expressions Using Topic Models and Intensity of Emotions}
\author{
Jing Peng \& Anna Feldman\\
  Computer Science/Linguistics \\
Montclair State University \\
  Montclair, New Jersey, USA\\ 
 {\tt \{pengj,feldmana\}@mail.montclair.edu}\\\And
 Ekaterina Vylomova \\
Computer Science \\
Bauman State Technical University\\
  Moscow, Russia\\
{\tt evylomova@gmail.com}
}

\date{}

\begin{document}
\maketitle
\begin{abstract}
We describe an algorithm for automatic classification of idiomatic and literal expressions. Our starting point is that words in a given text segment, such as a paragraph, that are high-ranking representatives of a common
topic of discussion are less likely to be a part of an idiomatic expression. Our additional hypothesis is that contexts in which idioms occur, typically, are more affective and therefore, we incorporate a simple analysis of the intensity of the emotions expressed by the contexts. We investigate the bag of words topic representation of one to three paragraphs containing an expression that should be classified as idiomatic or literal (a target phrase). We extract topics from paragraphs containing idioms and from paragraphs containing literals using an unsupervised
clustering method, Latent Dirichlet Allocation (LDA)
\citep{blei-etal:03}.  Since idiomatic expressions exhibit the property of non-compositionality, we assume that they usually present different semantics than the words used in the local topic. We treat idioms as semantic outliers, and the identification of a semantic shift as outlier detection. 
 Thus, this topic representation allows us to differentiate idioms from
literals using local semantic contexts.  Our results are encouraging.

\end{abstract}

\section{Introduction}

The definition of what is literal and figurative is still object of
debate. \cite{ariel:02} demonstrates that literal and non-literal
meanings cannot always be distinguished from each other. Literal
meaning is originally assumed to be conventional, compositional,
relatively context independent, and truth conditional. The problem is
that the boundary is not clear-cut, some figurative expressions are
compositional -- metaphors and many idioms; others are conventional --
most of the idioms. Idioms present great challenges for many Natural
Language Processing (NLP) applications. They can violate selection
restrictions \citep{sporleder-li:eacl09} as in \emph{push one's luck}
under the assumption that only concrete things can normally be
pushed. Idioms can disobey typical subcategorization constraints
(e.g., \emph{in line} without a determiner before line), or change the
default assignments of semantic roles to syntactic categories (e.g.,
in \emph{X breaks something with Y}, Y typically is an instrument but
for the idiom \emph{break the ice}, it is more likely to fill a
patient role as in \emph{How to break the ice with a stranger}). In
addition, many potentially idiomatic expressions can be used either
literally or figuratively, depending on the context.  This presents a
great challenge for machine translation. For example, a machine
translation system must translate \emph{held fire} differently in
\emph{Now, now, hold your fire until I've had a chance to
  explain. Hold your fire, Bill. You're too quick to complain.} and
\emph{The sergeant told the soldiers to hold their fire. Please hold
  your fire until I get out of the way}. In fact, we tested the last
two examples using the Google Translate engine and we got proper
translations of the two neither into Russian nor into Hebrew,
Spanish, or Chinese. Most current translation systems rely on large repositories
of idioms. Unfortunately, these systems are not capable to tell apart
literal from figurative usage of the same expression in context.  Despite the
common perception that phrases that can be idioms are mainly used in
their idiomatic sense, \cite{fazly-etal:09}'s analysis of 60 idioms
has shown that close to half of these also have a clear literal
meaning; and of those with a literal meaning, on average around 40\%
of their usages are literal.

In this paper we describe an algorithm for automatic classification of idiomatic and literal expressions. Our starting point is that words in a given text segment, such as a paragraph, that are high-ranking representatives of a common
topic of discussion are less likely to be a part of an idiomatic expression. Our additional hypothesis is that contexts in which idioms occur, typically,  are more affective and therefore, we incorporate a simple analysis of the intensity of the emotions expressed by the contexts. We investigate the bag of words \emph{topic} representation of one to three paragraphs containing an expression that should be classified as idiomatic or literal (a target phrase). We extract topics from paragraphs containing idioms and from paragraphs containing literals using an unsupervised
clustering method, Latent Dirichlet Allocation (LDA)
\citep{blei-etal:03}.  Since idiomatic expressions exhibit the property of non-compositionality, we assume that they usually present different semantics than the words used in the local topic. We treat idioms as semantic outliers, and the identification of semantic shift as outlier detection. 
 Thus, this topic representation allows us to differentiate idioms from
literals using the local semantics.  


The paper is organized as follows. Section \ref{sec:previous} briefly describes previous approaches to idiom recognition or classification.
In Section \ref{sec:approach} we describe our approach in detail, including the hypothesis, the topic space representation, and the proposed algorithm.
After describing the preprocessing procedure in Section \ref{sec:preprocessing}, we turn to the actual experiments in Sections \ref{sec:experiments} and \ref{sec:results}. We then compare our approach to other approaches (Section \ref{sec:other}) and discuss the results (Section \ref{sec:discuss}).

\section{Previous Work}\label{sec:previous}

Previous approaches to idiom detection can be classified into two
groups: 1) Type-based extraction, i.e., detecting idioms at the type
level; 2) token-based detection, i.e., detecting idioms in context.
Type-based extraction is based on the idea that idiomatic expressions
exhibit certain linguistic properties that can distinguish them from
literal expressions (\cite{sag-etal:02,fazly-etal:09}), among many
others, discuss various properties of idioms. Some examples of such
properties include 1) lexical fixedness: e.g., neither `shoot the
wind' nor `hit the breeze' are valid variations of the idiom shoot the
breeze and 2) syntactic fixedness: e.g., \emph{The guy kicked the
  bucket} is potentially idiomatic whereas \emph{The bucket was
  kicked} is not idiomatic anymore; and of course, 3)
non-compositionality.  Thus, some approaches look at the tendency for
words to occur in one particular order, or a fixed
pattern. \cite{hearst:92} identifies lexico-syntactic patterns that
occur frequently, are recognizable with little or no precoded
knowledge, and indicate the lexical relation of interest.
\cite{widdows-dorow:05} use Hearst's concept of lexicosyntactic
patterns to extract idioms that consist of fixed patterns between two
nouns. Basically, their technique works by finding patterns such as
``thrills and spills'', whose reversals (such as ``spills and
thrills'') are never encountered.

While many idioms do have these properties, many idioms fall on the
continuum from being compositional to being partly unanalyzable to
completely non-compositional
(\cite{cook-etal:07}). \cite{fazly-etal:09,li-sporleder:naacl10},
among others, notice that type-based approaches do not work on
expressions that can be interpreted idiomatically or literally
depending on the context and thus, an approach that considers tokens
in context is more appropriate for the task of idiom recognition.

A number of token-based approaches have been discussed in the
literature, both supervised (\cite{katz-giesbrecht:06}), weakly
supervised (\cite{birke-sarkar:06}) and unsupervised
(\cite{sporleder-li:eacl09,fazly-etal:09}).  \cite{fazly-etal:09}
develop statistical measures for each linguistic property of idiomatic
expressions and use them both in a type-based classification task and
in a token identification task, in which they distinguish idiomatic
and literal usages of potentially idiomatic expressions in context.
\cite{sporleder-li:eacl09} present a graph-based model for
representing the lexical cohesion of a discourse. Nodes represent
tokens in the discourse, which are connected by edges whose value is
determined by a semantic relatedness function. They experiment with
two different approaches to semantic relatedness: 1) Dependency
vectors, as described in \cite{pado-lapata:07}; 2) Normalized Google
Distance (\cite{cilibrasi-vitanyi:07}). \cite{sporleder-li:eacl09}
show that this method works better for larger contexts (greater than
five paragraphs). \cite{li-sporleder:naacl10} assume that literal and
figurative data are generated by two different Gaussians, literal and
non-literal and the detection is done by comparing which Gaussian
model has a higher probability to generate a specific instance. The
approach assumes that the target expressions are already known and the
goal is to determine whether this expression is literal or figurative
in a particular context. The important insight of this method is that
figurative language in general exhibits less semantic cohesive ties
with the context than literal language.

\cite{feldman-peng:13} describe several approaches to automatic idiom
identification. One of them is idiom recognition as outlier
detection. They apply principal component analysis for outlier
detection -- an approach that does not rely on costly annotated
training data and is not limited to a specific type of a syntactic
construction, and is generally language independent.  The quantitative
analysis provided in their work shows that the outlier detection
algorithm performs better and seems promising.  The qualitative
analysis also shows that their algorithm has to incorporate several
important properties of the idioms: \label{compositionality}(1) Idioms
are relatively non-compositional, comparing to literal expressions or
other types of collocations.
\label{topics}(2) Idioms violate local cohesive ties, as a result, they are
semantically distant from the local topics.  (3) While
not all semantic outliers are idioms, non-compositional semantic
outliers are likely to be idiomatic.  (4) Idiomaticity
is not a binary property. Idioms fall on the continuum from being
compositional to being partly unanalyzable to completely
non-compositional.

The approach described below is taking \cite{feldman-peng:13}'s
original idea and is trying to address (2) directly and (1)
indirectly.  Our approach is also somewhat similar to
\cite{li-sporleder:naacl10} because it also relies on a list of
potentially idiomatic expressions.

\section{Our Hypothesis}\label{sec:approach}

Similarly to \cite{feldman-peng:13}, out starting point is that idioms
are semantic outliers that violate cohesive structure, especially in
local contexts. However, our task is framed as supervised
classification and we rely on data annotated for idiomatic and literal
expressions.  We hypothesize that words in a given text segment, such
as a paragraph, that are high-ranking representatives of a common
topic of discussion are less likely to be a part of an idiomatic
expression in the document.

\subsection{Topic Space Representation}

Instead of the simple bag of words representation of a target document
(segment of three paragraphs that contains a target phrase), we
investigate the bag of words topic representation for target
documents. That is, we extract topics from paragraphs containing
idioms and from paragraphs containing literals using an unsupervised
clustering method, Latent Dirichlet Allocation (LDA)
\citep{blei-etal:03}.  The idea is that if the LDA model is able to
capture the semantics of a target document, an
idiomatic phrase will be a ``semantic'' outlier of the themes. Thus,
this topic representation will allow us to differentiate idioms from
literals using the semantics of the local context.

Let $d=\{w_1,\cdots,w_N\}^t$ be a segment (document) containing a
target phrase, where $N$ denotes the number of terms in a given
corpus, and $t$ represents transpose.  We first compute a set of $m$
topics from $d$. We denote this set by
\[
T(d)=\{t_1,\cdots,t_m\},
\]
where $t_i=(w_1,\cdots,w_k)^t$. Here $w_j$
represents a word from a vocabulary of $W$ words. Thus, we have two
representations for $d$: (1) $d$, represented by its original terms,
and (2) $\hat{d}$, represented by its topic terms. Two corresponding
term by document matrices will be denoted by $M_D$ and $M_{\hat{D}}$,
respectively, where $D$ denotes  a set of documents.  That is, $M_D$
represents the original ``text'' term by document matrix, while
$M_{\hat{D}}$ represents the ``topic'' term by document matrix.


\begin{figure*}[ht]
\centering
\epsfig{file=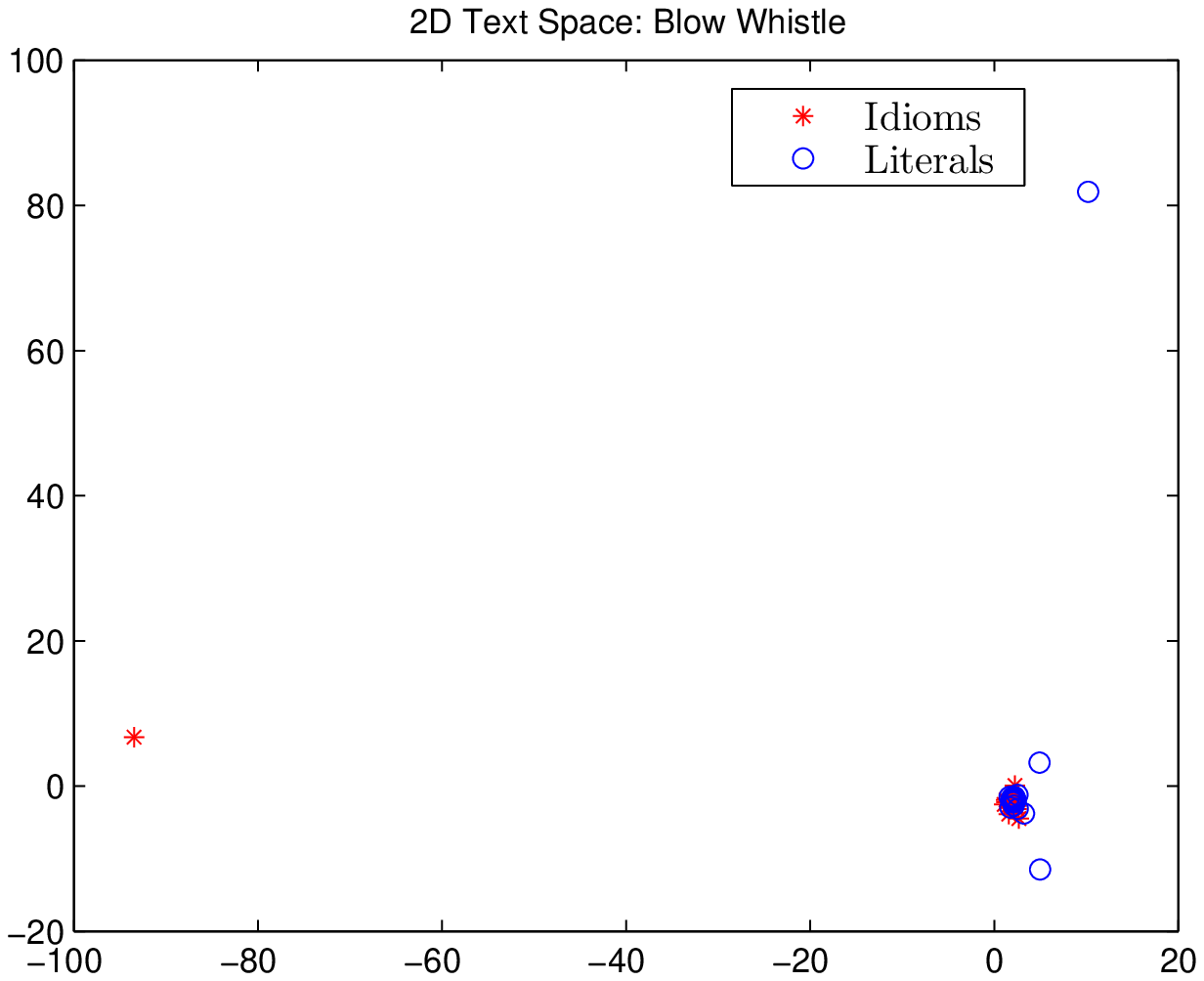,height=1.8in,width=2in}
\epsfig{file=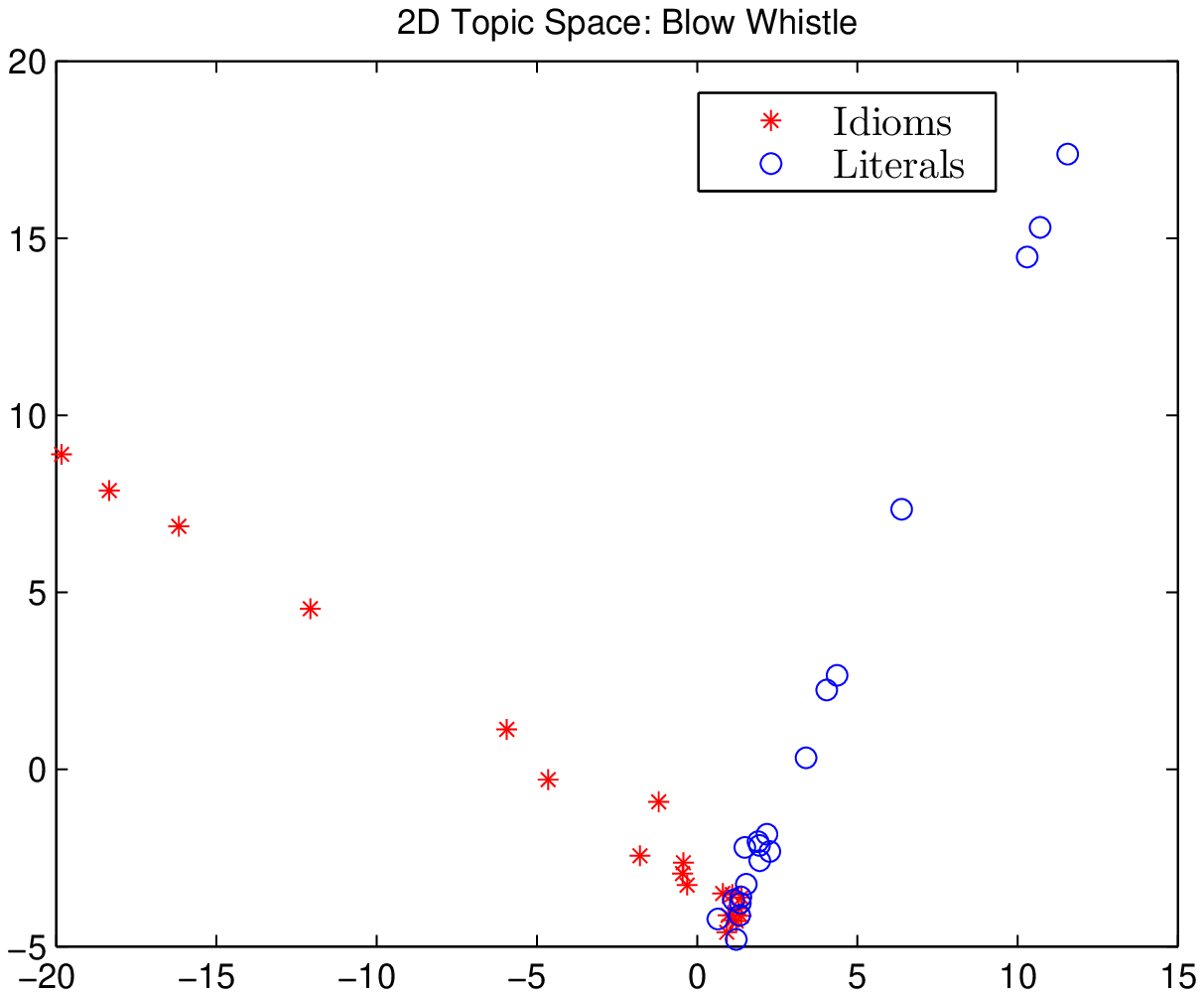,height=1.8in,width=2in}
\epsfig{file=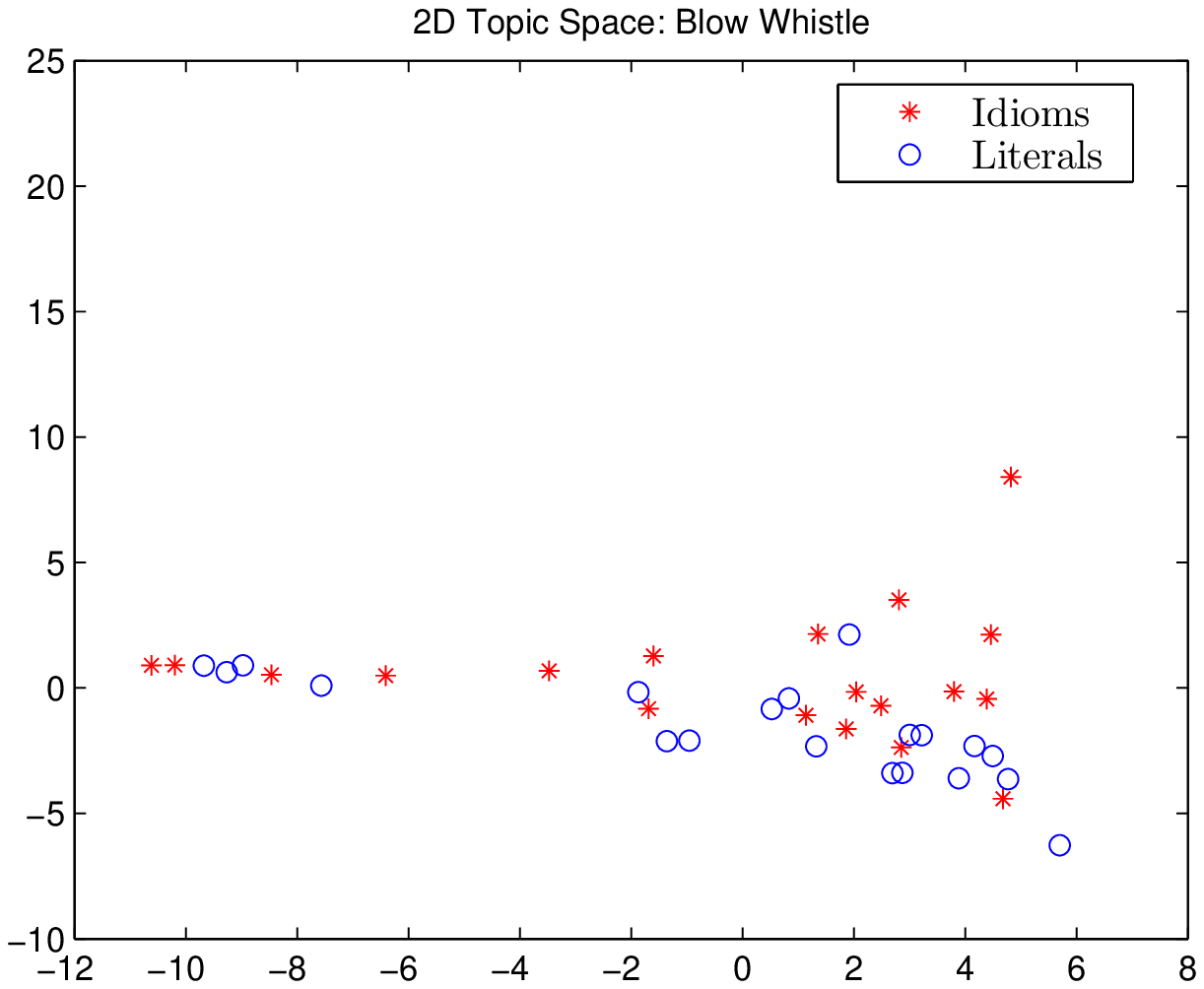,height=1.8in,width=2in}
\caption{2D projection of text segments containing ``blow whistle.''
  Left panel: Original text space. Middle panel: Topic space with
  restricted vocabulary. Right panel: Topic space with enlarged
  vocabulary.}
\label{TwoDProj}
\end{figure*}

Figure \ref{TwoDProj} shows the potential benefit of topic space
representation.  In the figure, text segments containing target phrase
``blow whistle'' are projected on a two dimensional subspace. The left
figure shows the projection in the ``text'' space, represented by the
term by document matrix $M_D$.  The middle figure shows the projection
in the topic space, represented by $M_{\hat{D}}$.  The topic space
representation seems to provide a better separation.

We note that when learning topics from a small data sample, learned
topics can be less coherent and interpretable, thus less useful.  To
address this issue, regularized LDA has been proposed in the
literature \citep{Newman:nips2011}.  A key feature is to favor words
that exhibit short range dependencies for a given topic. We can
achieve a similar effect by placing restrictions on the
vocabulary. For example, when extracting topics from segments
containing idioms, we may restrict the vocabulary to contain words
from these segments only. The middle and right figures in Figure
\ref{TwoDProj} illustrate a case in point. The middle figure shows a
projection onto the topic space that is computed with a restricted
vocabulary, while the right figure shows a projection when we place no
restriction on the vocabulary. That is, the vocabulary includes terms
from documents that contain both idioms and literals.

Note that by computing $M_{\hat{D}}$, the topic term by document
matrix, from the training data, we have created a vocabulary, or a set
of ``features'' (i.e., topic terms) that is used to directly describe
a query or test segment. The main advantage is that topics are more
accurate when computed by LDA from a large collection of idiomatic or literal contexts. 
Thus, these topics capture more accurately the  semantic contexts in which the target idiomatic and literal expressions typically occur. If a target query appears in a similar semantic context, the topics will be able to describe this query as well. On the other hand, one might similarly apply LDA to a given query to
extract query topics, and create the query vector from the query
topics.  The main disadvantage is that LDA may not be able to extract
topic terms that match well with those in the training corpus, when
applied to the query in isolation.

\subsection{Algorithm}

The main steps of the proposed algorithm, called {\bf TopSpace}, are
shown below.
\begin{algorithm}[ht]
\textbf{Input:} $D=\{d_1,\cdots,d_k,d_{k+1},\cdots,d_n\}$: 
training documents of $k$ idioms and $n-k$
literals. $Q=\{q_1,\cdots,q_l\}$: $l$ query documents.
\begin{enumerate}
\item
Let $DicI$ be the vocabulary determined solely from idioms
$\{d_1,\cdots,d_k\}$.  Similarly, let $DicL$ be the vocabulary
obtained from literals $\{d_{k+1},\cdots,d_n\}$.
\item
For a document $d_i$ in $\{d_1,\cdots,d_k\}$, apply LDA to extract a
set of $m$ topics $T(d_i)=\{t_1,\cdots,t_m\}$ using $DicI$. For $d_i
\in \{d_{k+1},\cdots,d_n\}$, $DicL$ is used.
\item
Let
$\hat{D}=\{\hat{d}_1,\cdots,\hat{d}_k,\hat{d}_{k+1},\cdots,\hat{d}_n\}$
be the resulting topic representation of $D$.
\item
Compute the term by document matrix $M_{\hat{D}}$ from $\hat{D}$, and
let $DicT$ and $gw$ be the resulting dictionary and global weight
($idf$), respectively.
\item
Compute the term by document matrix $M_{Q}$ from $Q$, using $DicT$ and
$gw$ from the previous step.
\end{enumerate}
{\bf Output}: $M_{\hat{D}}$ and $M_{Q}$
\label{Algorithm1}
\end{algorithm}

To summarize, after splitting our corpus (see section
\ref{sec:preprocessing}) into paragraphs and preprocessing it, we
extract topics from paragraphs containing idioms and from paragraphs
containing literals.  We then compute a term by document matrix, where
terms are topic terms and documents are topics extracted from the
paragraphs. Our test data are represented as a term-by-document matrix
as well (See the details in section \ref{sec:experiments}). 

\subsection{Fisher Linear Discriminant Analysis}

Once $M_{\hat{D}}$ and $M_Q$ are obtained, a classification rule can
be applied to predict idioms vs. literals.  The approach we are taking
in this work for classifying idioms vs.  literals is based on Fisher's
discriminant analysis (FDA) \citep{Fukunaga90}.  FDA often
significantly simplifies tasks such as regression and classification
by computing low-dimensional subspaces having statistically
uncorrelated or discriminant variables. In language analysis,
statistically uncorrelate or discriminant variables are extracted and
utilized for description, detection, and classification.
\cite{woods-etal:86}, for example, use statistically uncorrelated
variables for language test scores.  A group of subjects is scored on
a battery of language tests, where the subtests measure different
abilities such as vocabulary, grammar or reading
comprehension. \cite{horvath:85} analyzes speech samples of Sydney
speakers to determine the relative occurrence of five different
variants of each of five vowels sounds. Using this data, the speakers
cluster according to such factors as gender, age, ethnicity and
socio-economic class.

A similar approach has been discussed in
\cite{peng-feldman-street:10}.  FDA is a class of methods used in
machine learning to find the linear combination of features that best
separate two classes of events.  FDA is closely related to principal
component analysis (PCA), where a linear combination of features that
best explains the data. Discriminant analysis explicitly exploits
class information in the data, while PCA does not.

Idiom classification based on discriminant analysis has several
advantages.  First, as has been mentioned, it does not make any
assumption regarding data distributions.  Many statistical detection
methods assume a Gaussian distribution of normal data, which is far
from reality.  Second, by using a few discriminants to describe data,
discriminant analysis provides a compact representation of the data,
resulting in increased computational efficiency and real time
performance.

In FDA, within-class, between-class, and mixture scatter matrices are
used to formulate the criteria of class separability. Consider a $J$
class problem, where $m_{0}$ is the mean vector of all data, and
$m_{j}$ is the mean vector of $j$th class data. A within-class scatter
matrix characterizes the scatter of samples around their respective
class mean vector, and it is expressed by
\begin{equation}
S_{w}=\sum_{j=1}^{J}p_j\sum_{i=1}^{l_{j}}(x_{i}^{j}-m_{j})(x_{i}^{j}-m_{j})^t,
\end{equation}
where $l_{j}$ is the size of the data in the $j$th class, $p_j$
($\sum_jp_j=1$) represents the proportion of the $j$th class
contribution, and $t$ denotes the transpose operator.  A between-class
scatter matrix characterizes the scatter of the class means around the
mixture mean $m_{0}$. It is expressed by
\begin{equation}
S_{b}=\sum_{j=1}^{J}p_j(m_{j}-m_{0})(m_{j}-m_{0})^t.
\end{equation}
The mixture scatter matrix is the covariance matrix of all samples,
regardless of their class assignment, and it is given by
\begin{equation}
S_{m}=\sum_{i=1}^{l}(x_{i}-m_{0})(x_{i}-m_{0})^t =S_{w}+S_{b}.
\label{total}
\end{equation}
The Fisher criterion is used to find a projection matrix $W \in
\Re^{q\times d}$ that maximizes
\begin{equation}
J(W)=\frac{|W^tS_{b}W|}{|W^tS_{w}W|}.
\label{Fisher_criterion}
\end{equation}
In order to determine the matrix $W$ that maximizes $J(W)$, one can
solve the generalized eigenvalue problem: $S_{b}w_{i}=\lambda
_{i}S_{w}w_{i}$. The eigenvectors corresponding to the largest
eigenvalues form the columns of $W$.  For a two class problem, it can
be written in a simpler form: $S_{w}w=m=m_{1}-m_{2}$, where $m_{1}$
and $m_{2}$ are the means of the two classes.

\section{Data preprocessing}
\label{sec:preprocessing}

\subsection{Verb-noun constructions}

For our experiments we use the British National Corpus (BNC,
\cite{burnard:00}) and a list of verb-noun constructions (VNCs)
extracted from BNC by \cite{fazly-etal:09,Cook2008} and labeled as L
(Literal), I (Idioms), or Q (Unknown).  The list contains only those
VNCs whose frequency was greater than 20 and that occurred at least in
one of two idiom dictionaries
\citep{cowie-etal:83,seaton-macaulay:02}.  The dataset consists of
2,984 VNC tokens. For our experiments we only use VNCs that are
annotated as I or L.

\subsection{Lemmatization}

Instead of dealing with various forms of the same root, we use lemmas
provided by the BNC XML annotation, so our corpus is lemmatized. We
also apply the (modified) Google stop list before extracting the
topics. The reason we modified the stop list is that some function
words can potentially be idiom components (e.g., certain
prepositions).


\subsection{Paragraphs}

We use the original SGML annotation to extract paragraghs from BNC. We
only kept the paragraphs that contained VNCs for our experiments.  We
experimented with texts of one paragraph length (single paragraph
contexts) and of three-paragraph length (multi-paragraph contexts).  
An example of multi-paragraph contexts is shown below:
\bigskip
{\em 

So, reluctantly, I joined Jack Hobbs in not rocking the boat, reporting the play and the general uproar with perhaps too much impartiality.
My reports went to all British newspapers, with special direct services by me to India, South Africa and West Indies; even to King George V in Buckingham Palace, who loved his cricket. In other words, I was to some extent leading the British public astray.

I regret I can shed little new light on the mystery of who {\bf blew the whistle} on the celebrated dressing-room scene after Woodfull was hit.
while he was lying on the massage table after his innings waiting for a doctor, Warner and Palairet called to express sympathy.

Most versions of Woodfull's reply seem to agree that he said. There are two teams out there on the oval.
 One is playing cricket, the other is not.
This game is too good to be spoilt.
It is time some people got out of it.
Warner and Palairet were too taken aback to reply.
They left the room in embarrassment.
}

Single paragraph contexts simply consist of the middle paragraph.

\section{Experiments}
\label{sec:experiments}

\subsection{Methods}

We have carried out an empirical study evaluating the performance of
the proposed algorithm. For comparison, the following methods are
evaluated. (1) The proposed algorithm {\bf TopSpace}
(\ref{Algorithm1}), where the data are represented in topic space. (2)
{\bf TexSpace} algorithm, where the data are represented in original
text space.  For each representation, two classification schemes are
applied: a) FDA (Eq. \ref{Fisher_criterion}), followed by the nearest
neighbor rule. b) SVMs with Gaussian kernels
(\cite{cristianini:introSVM}).  For the nearest neighbor rule, the
number of nearest neighbors is set to $\lceil n/5\rceil$, where $n$
denotes the number of training examples. For SVMs, kernel width and
soft margin parameters are set to default values.

\begin{table*}
\caption{Average accuracy of competing methods on four datasets in single paragraph
  contexts: A = Arousal}
\label{single}	
\begin{center}
\setlength{\tabcolsep}{.8ex}
\begin{tabular}{|l|ccc|ccc|ccc|ccc|}
\hline
{\bf Model} & \multicolumn{3}{c}{\bf BlowWhistle}& \multicolumn{3}{c|}{\bf LoseHead}		&		\multicolumn{3}{c|}{\bf MakeScene}	&			\multicolumn{3}{c|}{\bf TakeHeart}\\
\hline	
	& Prec & 	Recall & 	Acc & 		Prec & 	Recall & 	Acc	& 	Prec & 	Recall & 	Acc & 		Prec & 	Recall & 	Acc\\
	\hline
FDA-Topics &  0.44&0.40&0.79&{\bf 0.70}&{\bf 0.90}&{\bf 0.70}&{\bf 0.82}&{\bf 0.97}&{\bf 0.81}&0.91&0.97&0.89\\
FDA-Topics+A& 0.51&0.51&0.75&0.78&0.68&0.66&0.80&0.99&0.80&0.93&0.84&0.80\\
FDA-Text   &  {\bf 0.37}&{\bf 0.81}&{\bf 0.63}&0.60&0.88&0.58&0.82&0.89&0.77&0.36&0.38&0.41\\
FDA-Text+A &  0.42&0.49&0.76&0.64&0.92&0.63&{\bf 0.83}&{\bf 0.95}&{\bf 0.82}&0.75&0.53&0.53\\
SVMs-Topics&  0.08&0.39&0.59&0.28&0.25&0.45&0.59&0.74&0.61&{\bf 0.91}&{\bf 1.00}&{\bf 0.91}\\
SVMs-Topics+A&0.06&0.21&0.69      &0.38      &0.18      &0.44      &0.53      &0.40      &0.44     &{\bf 0.91}      &{\bf 1.00}&{\bf 0.91}\\
SVMs-Text     &0.08      &0.39      &0.59      &0.36      &0.60      &0.52      &0.23      &0.30     &0.40     &0.42      &0.16      &0.22\\
SVMs-Text+A  &0.15      &0.51      &0.60      &0.31      &0.38      &0.48      &0.37      &0.40      &0.45     &0.95&0.48      &0.50\\ \hline
\end{tabular}
\end{center}	
\end{table*}

\subsection{Data Sets}

The following data sets are used to evaluate the performance of the
proposed technique. These data sets have enough examples from both
idioms and literals to make our results meaningful. On average, the
training data is 6K word tokens.  Our test data is of a similar size.

{\bf BlowWhistle}: This data set has 78 examples, 27 of which are
idioms and the remaining 51 are literals. The training data for {\bf
  BlowWhistle} consist of 40 randomly chosen examples (20 paragraphs
containing idioms and 20 paragraphs containing literals). The
remaining 38 examples (7 idiomatic and 31 literals) are used as test
data.

{\bf MakeScene}: This data set has 50 examples, 30 of which are
paragraphs containing idioms and the remaining 20 are paragraphs
containing literals. The training data for {\bf MakeScene} consist of
30 randomly chosen examples, 15 of which are paragraphs containing
\emph{make scene} as an idiom and the rest 15 are paragraphs
containing \emph{make scene} as a literal. The remaining 20 examples
(15 idiomatic paragraphs and 5 literals) are used as test data.

{\bf LoseHead}: This data set has 40 examples, 21 of which are idioms
and the remaining 19 are literals. The training data for {\bf
  LoseHead} consist of 30 randomly chosen examples (15 idiomatic and
15 literal). The remaining 10 examples (6 idiomatic and 4 literal) are
used as test data.

{\bf TakeHeart}: This data set has 81 examples, 61 of which are idioms
and the remaining 20 are literals. The training data for {\bf
  TakeHeart} consist of 30 randomly chosen examples (15 idiomatic and
15 literals). The remaining 51 examples (46 idiomatic and 5 literals)
are used as test data.

\subsection{Adding affect}

\cite{nunberg-etal:94} notice that \lq\lq{}idioms are typically used
to imply a certain evaluation or affective stance toward the things
they denote\rq\rq{}. Language users usually choose an idiom in
non-neutral contexts. The situations that idioms describe can be
positive or negative; however, the polarity of the context is not as
important as the strength of the emotion expressed. So, we decided to
incorporate the knowledge about the emotion strength into our
algorithm.  We use a database of word norms collected by
\cite{warriner-etal:13}. This database contains almost 14,000 English
lemmas annotated with three components of emotions: valence (the
pleasantness of a stimulus), arousal (the intensity of emotion
provoked by a stimulus), and dominance (the degree of control exerted
by a stimulus). These components were elicited from human subjects via
an Amazon Mechanical Turk crowdsourced experiment. We only used the
arousal feature in our experiments because we were interested in the
intensity of the emotion rather than its valence.

For a document $d=\{w_1,\cdots,w_N\}^t$, we calculate the
corresponding arousal value $a_i$ for each $w_i$, obtaining
$d_{A}=\{a_1,\cdots,a_N\}^t$. Let $m_{A}$ be the average arousal value
calculated over the entire training data. The centered arousal value
for a training document is obtained by subtracting $m_{A}$ from
$d_{A}$, i.e.,
$\bar{d}_{A}=d_{A}-m_{A}=\{a_1-m_{A},\cdots,a_N-m_{A}\}^t$.
Similarly, the centered arousal value for a query is computed
according to $\bar{q}_{A}=q_{A}-m_{A}=\{q_1-m_{A},\cdots,q_N-m_{A}\}^t$. That is,
the training arousal mean is used to center both training and query
arousal values.  The corresponding arousal matrices for $D$,
$\hat{D}$, and $Q$ are $A_{D}$, $A_{\hat{D}}$, $A_Q$, respectively. To
incorporate the arousal feature, we simply compute
\begin{equation}
\Theta_D=M_D+A_D, 
\label{textA}
\end{equation}
and
\begin{equation}
\Theta_{\hat{D}}=M_{\hat{D}}+A_{\hat{D}}.
\label{topicA}
\end{equation}
The arousal feature can be similarly incoporated into query
$\Theta_Q=M_Q+A_Q$.

\section{Results}\label{sec:results} 

Table \ref{single} shows the average precision, recall, and accuracy
of the competing methods on the four data sets over 10 runs in simple
paragraph contexts. Table \ref{multi} shows the results for the
multi-paragraph contexts. Note that for single paragraph contexts, we
chose two topics, each having 10 terms. For multi-paragrah contexts,
we had four topics, with 10 terms per topic.  No optimization was made
for selecting the number of topics as well as the number of terms per
topic.  In the tables, the best performance in terms of the sum of
precision, recall and accuracy is given in boldface.

\begin{table*}
\caption{Average accuracy of competing methods on four datasets
  in multiple paragraph contexts: A = Arousal}
\label{multi}	
\begin{center}
\setlength{\tabcolsep}{.8ex}
\begin{tabular}{|l|ccc|ccc|ccc|ccc|}
\hline
{\bf Model} & \multicolumn{3}{c}{\bf BlowWhistle}& \multicolumn{3}{c|}{\bf LoseHead}		&		\multicolumn{3}{c|}{\bf MakeScene}	&			\multicolumn{3}{c|}{\bf TakeHeart}\\
\hline	
	& Prec & 	Recall & 	Acc & 		Prec & 	Recall & 	Acc	& 	Prec & 	Recall & 	Acc & 		Prec & 	Recall & 	Acc\\
	\hline
FDA-Topics  				 &{\bf 0.62}&{\bf 0.60} &{\bf 0.83} &{\bf 0.76}&{\bf 0.97}&{\bf 0.78}&0.79 &0.95 &0.77 &{\bf 0.93}&{\bf 0.99}&{\bf 0.92}\\
FDA-Topics+A 					&0.47 &        0.44 &         0.79 &0.74 &0.93 &0.74 &0.82&0.69 &0.65 &0.92 &0.98&0.91\\
FDA-Text     					 &0.65 &0.43 &0.84&0.72 &0.73 &0.65 &0.79 &0.95 &0.77 &0.46 &0.40&0.42\\
FDA-Text+A  				          &0.45 &0.49 &0.78 &0.67 &0.88 &0.65 &{\bf 0.80} &{\bf 0.99}&{\bf 0.80}&0.47 &0.29&0.33\\
SVMs-Topics  					&0.07 &0.40 &0.56 &0.60 &0.83 &0.61 &0.46 &0.57 &0.55 &0.90 &1.00&0.90\\
SVMs-Topics+A                                     &0.21 &0.54 &0.55 &0.66 &0.77 &0.64 &0.42 &0.29 &0.41 &0.91 &1.00&0.91\\
SVMs-Text                                             &0.17 &0.90&0.25 &0.30 &0.50 &0.50 &0.10 &0.01 &0.26 &0.65 &0.21&0.26\\
SVMs-Text+A  &	                          0.24 &0.87 &0.41 &0.66 &0.85 &0.61 &0.07 &0.01 &0.26 &0.74 &0.13&0.20\\ \hline
\end{tabular}
\end{center}	
\end{table*}

\begin{figure}[ht]
\centering
\epsfig{file=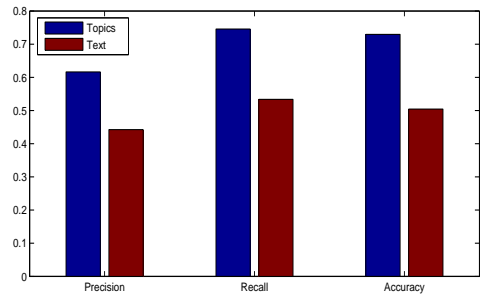,height=1.7in,width=3in}
\caption{Aggregated performance: Topic vs text representations.}
\label{TopTex}
\end{figure}

The results show that the topic representation achieved the best
performance in 6 out of 8 cases.  Figure \ref{TopTex} plots the
overall aggregated performance in terms of topic vs text
representations across the entire data sets, regardless of the
classifiers used.  Everything else being equal, this clearly shows the
advantage of topics over simple text representation.

\begin{figure}[ht]
\centering
\epsfig{file=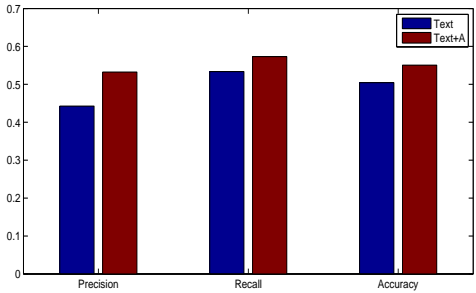,height=1.7in,width=3in}\\
\epsfig{file=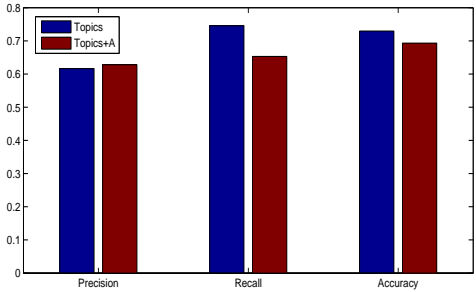,height=1.7in,width=3in}
\caption{Aggregated performance: Text vs.~ text+Arousal representations
  (top) and Topics vs.~ Topics+Arousal representations (bottom).}
\label{TTA}
\end{figure}

The arousal feature (Eqs \ref{textA} and \ref{topicA}) also improved the 
overall performance, particularly in text representation
(Eq. \ref{textA}). This can be seen in the top panel in Figure
\ref{TTA}. In fact, in 2/8 cases, text representation coupled with the
arousal feature achieved the best performance. One possible
explanation is that the LDA model already performed ``feature''
selection (choosing topic terms), to the extent possible. Thus, any
additional information such as arousal only provides marginal
improvement at the best (bottom panel in Figure \ref{TTA}). On the
other hand, original text represents ``raw'' features, whereby arousal
information helps provide better contexts, thus improving overall
performance.

\begin{figure}[ht]
\centering
\epsfig{file=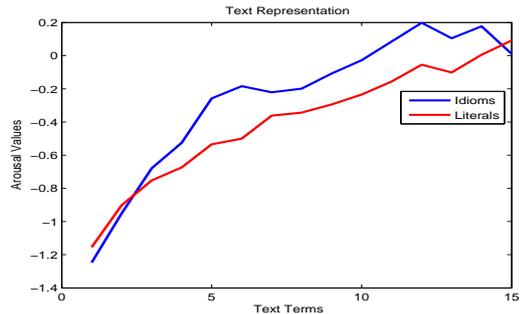,height=1.7in,width=3in}\\
\epsfig{file=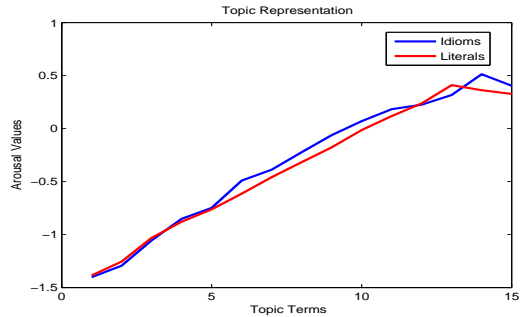,height=1.7in,width=3in}
\caption{Average arousal values--Upper panel: Text space. Lower panel:
  Topic space.}
\label{Av}
\end{figure}

Figure \ref{Av} shows a case in point: the average (sorted) arousal
values of idioms and literals of the target phrase ``lose head.'' The
upper panel plots arousal values in the text space, while lower panel
plots arousal values in the topic space. The plot supports the results
shown in Tables \ref{single} and \ref{multi}, where the arousal
feature generally improves text representation.

\section{Comparisons with other approaches}
\label{sec:other}

Even though we used \cite{fazly-etal:09}'s dataset for these
experiments, the direct comparison with their method is impossible
here because our task is formulated differently and we do not use the
full dataset for the experiments. \cite{fazly-etal:09}'s unsupervised
model that relies on the so-called canonical forms gives
72.4\% (macro-)accuracy on the extraction of idiomatic tokens when
evaluated on their test data.

We cannot compare our method directly with the other methods discussed
in section \ref{sec:previous} either because each uses a different
dataset or formulates the task differently (detection vs. recognition
vs. identification). However, we can compare the method presented here
with \cite{feldman-peng:13} who also experiment with LDA, use similar
data, and frame the problem as classification. Their goal, however, is
to classify \emph{sentences} as either idiomatic or literal.  To
obtain a discriminant subspace, they train their model on a small
number of randomly selected idiomatic and non-idiomatic
sentences. They then project both the training and the test data on
the chosen subspace and use the three nearest neighbor (3NN)
classifier to obtain accuracy.  The average accuracy they report is
80\%.  Our method clearly outperforms the \cite{feldman-peng:13}
approach (at least on the dataset we use).

\section{Discussion and Conclusion}
\label{sec:discuss} 

We have described an algorithm for automatic classification of
idiomatic and literal expressions.  We have investigated the bag of
words topic representation for target documents (segments of one or
three paragraphs that contains a target phrase). The approach
definitely outperforms the baseline model that is based on the simple
bag of words representation, but it also outperforms approaches
previously discussed in the literature.  Our model captures the local
semantics and thus is capable to identify semantic outliers (=idioms).

While we realize that the data set we use is small, the
results are encouraging.  We notice that using 3 paragraphs for local
contexts improves the performance of the classifiers. The reason is
that some paragraphs are relatively short. A larger context provides
more related terms, which gives LDA more opportunities to sample these
terms.

Idioms are also relatively non-compositional. While we do not measure
their non-compositionality in this approach, we indirectly touch upon
this property by hypothesizing that non-compositional idiomatic
expressions are likely to be far from the local topics.

We feel that incorporating the intensity of emotion expressed by the
context into our model improves performance, in particular, in text
representation. When we performed a qualitative analysis of the results trying to determine the causes of false positives and negatives, we noticed that there were quite a number of cases that improved after incorporating  the arousal feature into the model.
For example, the FDA:topic classifier labels "blow the whistle" as literal in the following context, but FDA:topics+A marks this expression as idiomatic (italicized words indicate words with relatively high arousal values):

\smallskip
\begin{small}
Peter thought it all out very \emph{carefully}.
He decided the \emph{ wisest} course was to pool all he had made over the last two years, enabling Julian to purchase the lease of a high street property. This would enable them to set up a business on a more \emph{settled} and \emph{permanent} trading basis. Before long they opened a grocery-cum-delicatessen in a \emph{good} position as far as passing trade was concerned. Peter's investment was not misplaced. The business did very well with the two lads greatly \emph{appreciated} locally for their \emph{hard} work and quality of service. The range of goods they were able to carry was \emph{welcomed} in the area, as well as lunchtime sandwich facilities which had previously been missing in the neighbourhood.

\smallskip
\emph{Success} was the fruit of some three years' \emph{strenuous} work.
But it was more than a \emph{shock} when Julian admitted to Peter that he had been running up \emph{huge debts} with 
their bank.
Peter knew that Julian \emph{gambled}, but he hadn't expected him to \emph{gamble} to that level, and certainly not to use the shop as security.
With continual borrowing over two years, the bank had \textbf{blown the whistle}.
Everything was gone.
Julian was \emph{bankrupt.}
Even if they'd had a formal partnership, which they didn't, it would have made no difference.
Peter \emph{lost} all he'd made, and with it his \emph{chance} to help his parents and his younger brother and sister, Toby and Laura.

\smallskip
Peter was \emph{heartbroken}.
His father had said all along: neither a lender nor a borrower.
Peter had found out the \emph{hard} way.
But as his mother observed, he was the same Peter, he'd pick himself up somehow.
Once again, Peter was resolute.
He made up his mind he'd never make the same \emph{mistake} twice.
It wasn't just the money or the hard work, though the waste of that was \emph{difficult} enough to accept.
Peter had been working a \emph{debt} of \emph{love}.
He'd done all this for his parents, particularly for his father, whose \emph{dedication} to his children had always \emph{impressed} Peter and moved him \emph{deeply}.
And now it had all come to nothing.

\end{small}

Therefore, we think that idioms have the tendency to appear in  more affective contexts; and we think that incorporating more sophisticated sentiment analysis into our model will improve the results.


\section*{Acknowledgments}
This material is based upon work supported by the National Science Foundation under Grant No. 1319846.
We also thank the anonymous reviewers for useful comments.
The third author thanks the Fulbright Foundation for giving her an opportunity to conduct this research at Montclair State University (MSU). 


\bibliographystyle{chicago}

\bibliography{figurative,outlier}

\end{document}